\documentclass[10pt,twocolumn,letterpaper]{article}

\usepackage{cvpr}
\usepackage{times}
\usepackage{epsfig}
\usepackage{graphicx}
\usepackage{amsmath}
\usepackage{amssymb}
\DeclareMathOperator*{\argmin}{argmin}
\usepackage{gensymb}
\usepackage{pifont}
\usepackage{color}
\newcommand{\yj}{\textcolor{blue}}

\newcommand{\ignore}[1]{}
\usepackage[pagebackref=true,breaklinks=true,letterpaper=true,colorlinks,bookmarks=false]{hyperref}
\usepackage{cleveref}

\cvprfinalcopy %

\def\cvprPaperID{3171} %

\ifcvprfinal\fi
\begin{document}

\title{Interspecies Knowledge Transfer for Facial Keypoint Detection}%

\author{Maheen Rashid\\
University of California, Davis\\
{\tt\small mhnrashid@ucdavis.edu}
\and
Xiuye Gu\thanks{Work done while an intern at UC Davis.}\\
Zhejiang University\\
{\tt\small gxy0922@zju.edu.cn}
\and
Yong Jae Lee\\
University of California, Davis\\
{\tt\small yongjaelee@ucdavis.edu}
}
\maketitle
\begin{abstract}
We present a method for localizing facial keypoints on animals by transferring knowledge gained from human faces. Instead of directly finetuning a network trained to detect keypoints on human faces to animal faces (which is sub-optimal since human and animal faces can look quite different), we propose to first adapt the animal images to the pre-trained human detection network by correcting for the differences in animal and human face shape. We first find the nearest human neighbors for each animal image using an unsupervised shape matching method. We use these matches to train a thin plate spline warping network to warp each animal face to look more human-like. The warping network is then jointly finetuned with a pre-trained human facial keypoint detection network using an animal dataset. We demonstrate state-of-the-art results on both horse and sheep facial keypoint detection, and significant improvement over simple finetuning, especially when training data is scarce.  Additionally, we present a new dataset with 3717 images with horse face and facial keypoint annotations.
\end{abstract}

\section{Introduction}

Facial keypoint detection is a necessary precondition for face alignment and registration, and impacts facial expression analysis, facial tracking, as well as graphics methods that manipulate or transform faces. While human facial keypoint detection is a mature area of research, despite its importance, animal facial keypoint detection is a relatively unexplored area.  For example, veterinary research has shown that horses~\cite{gleerup2015equine,dalla2014development}, mice~\cite{langford2010coding}, sheep~\cite{boissy2011cognitive}, and cats~\cite{holden2014evaluation} display facial expressions of pain -- a facial keypoint detector could be used to help automate such animal pain detection.  In this paper, we tackle the problem of facial keypoint detection for animals, with a focus on horses and sheep.

Convolutional neural networks (CNNs) have demonstrated impressive performance for \emph{human} facial keypoint detection~\cite{peng2016recurrent,xiao2016robust,trigeorgis2016mnemonic,zhang2016occlusion,jourabloo2016large,zhu2016face,chen2016supervised,zhang2016joint}, which makes CNNs an attractive choice for learning facial keypoints on animals.  Unfortunately, training a CNN from scratch typically requires large amounts of labeled data, which can be time-consuming and expensive to collect.  Furthermore, while a CNN can be finetuned when there is not enough training data for the target task, a pre-trained network's extent of learning is limited both by the amount of data available for fine-tuning, as well as the \emph{relatedness of the two tasks}.  For example, previous work demonstrate that a network trained on man-made objects has limited ability to adapt to natural objects~\cite{yosinski2014transferable}, and additional pretraining data is only beneficial when related to the target task~\cite{huh2016makes}.

\begin{figure}[t]
	\centering
	\includegraphics[width=0.47\textwidth]{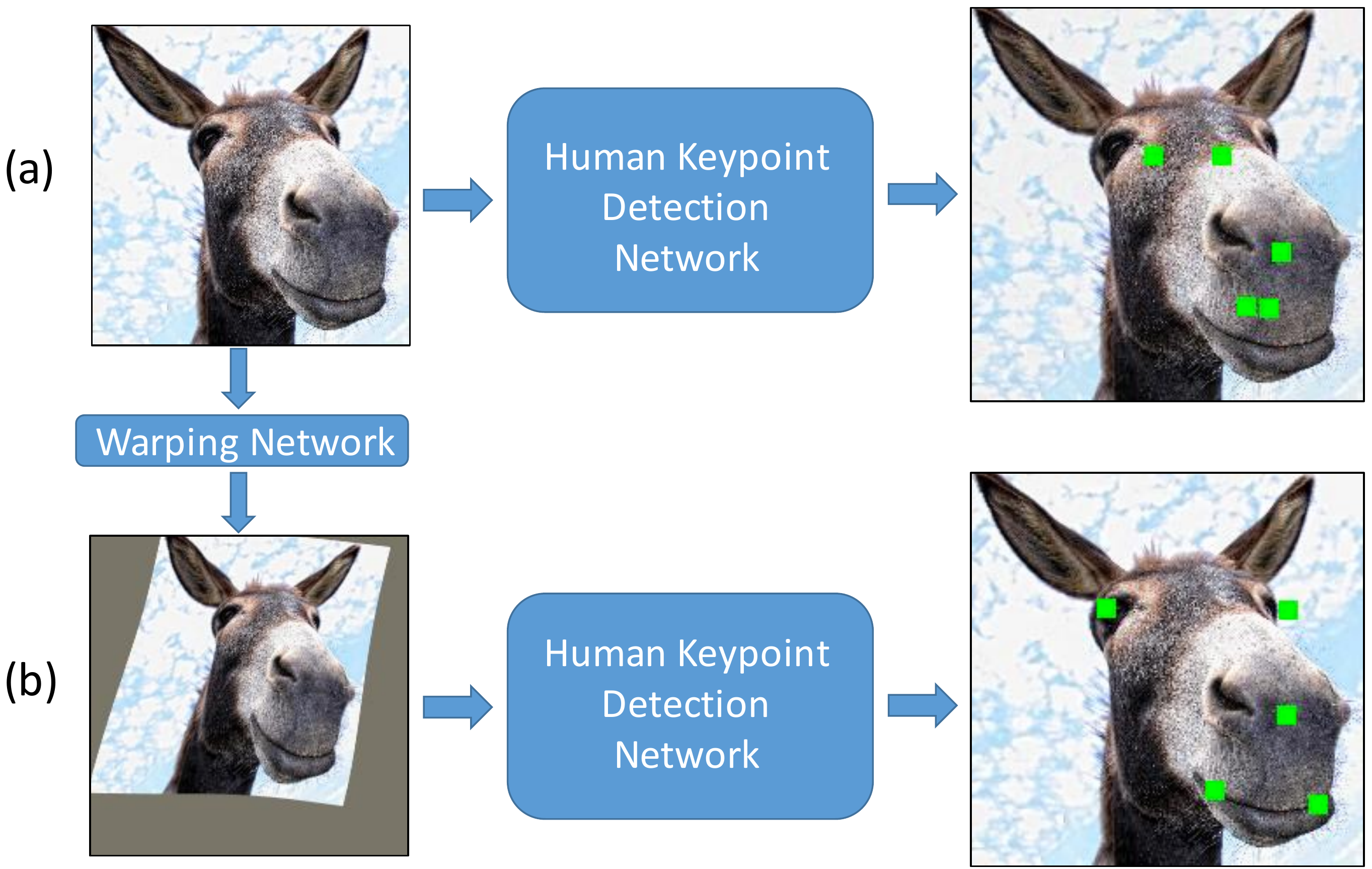}
	\caption{\textbf{Main idea.}  (\textbf{a}) Directly finetuning a human keypoint detector to horses can be suboptimal, since horses and humans have very different shapes and appearances. (\textbf{b})  By warping a horse to have a more human-like shape, the pre-trained human keypoint detector can more easily adapt to the horse's appearance.}
	\label{fig:concept}
\end{figure}

While there are large datasets with human facial keypoint annotations (e.g., AFLW has $\sim$26000 images~\cite{koestinger2011annotated}), there are, unfortunately, no large datasets of animal facial keypoints that could be used to train a CNN from scratch (e.g., the sheep dataset from \cite{yang2015human} has only $\sim$600 images).  At the same time, the structural differences between a human face and an animal face means that directly fine-tuning a human keypoint detector to animals can lead to a sub-optimal solution (as we demonstrate in Sec.~\ref{experiments}).

In this paper, we address the problem of transferring knowledge between two different types of data (human and animal faces) for the same task (keypoint detection). How can we achieve this with a CNN?  Our key insight is that rather than adapt a pre-trained network to training data in a new domain, we can first do the \emph{opposite}.  That is, \emph{we can adapt the training data from the new domain to the pre-trained network}, so that it is better conditioned for finetuning.  By mapping the new data to a distribution that better aligns with the data from the pre-trained task, we can take a pre-trained network from the loosely-related task of human facial keypoint detection and finetune it for animal facial keypoint detection.  Specifically, our idea is to explicitly warp each animal image to look more human-like\ignore{(using a TPS transformation)}, and then use the resulting warped images to finetune a network pre-trained to detect human facial keypoints.  See Fig.~\ref{fig:concept}.

Intuitively, by warping animal faces to look more human-like we can correct for their shape differences, so that during finetuning the network need only adapt to their differences in appearance. For example, the distance between the corners of a horse's mouth is typically much smaller than the distance between its eyes, whereas for a human these distances are roughly similar -- a shape difference. In addition, horses have fur, and humans do not -- an appearance difference. Our warping network adjusts for the shape difference by stretching out the horse's mouth corners, while during finetuning the keypoint detection network learns to adjust for the appearance difference.%

\vspace{-10pt}
\paragraph{Contributions.} Our contributions are three fold: First, we introduce a novel approach for animal facial keypoint detection that transfers knowledge from the loosely-related domain of human facial keypoint detection.\ignore{ We achieve this by reducing the discrepancy in shape between the animal and human faces via a warp transformation.} Second, we provide a new annotated horse facial keypoint dataset consisting of 3717 images. Third, we demonstrate state-of-the-art results on keypoint detection for horses and sheep.  By transforming the animal data to look more human-like, we attain significant gains in keypoint detection accuracy over simple finetuning. Importantly, the gap between our approach and simple finetuning widens as the amount of training data is reduced, which shows the practical applicability of our approach to small datasets. Our data and code are available at \url{https://github.com/menoRashid/animal_human_kp}. 
\section{Related work}

Facial landmark detection and alignment are mature topics of research in computer vision. Classic approaches include Active Appearance Models~\cite{cootes2001active,matthews2004active,saragih2007nonlinear,tzimiropoulos2013optimization}, Constrained Local Models~\cite{cristinacce2006feature,cristinacce2008automatic,saragih2011deformable,asthana2013robust}, regression based methods~\cite{valstar2010facial,xiong2013supervised,cao2014face,xiong2015global} with a cascade~\cite{dollar2010cascaded,lee2015face,zhu2015face}, and an ensemble of exemplar based models~\cite{belhumeur2013localizing}.  Recent work extends cascaded regression models by learning predictions from multiple domain-specific regressors~\cite{zhu2016unconstrained} or by using a mixture of regression experts at each cascade level~\cite{tuzel2016robust}. These models also demonstrate good performance when solved simultaneously with a closely related task, such as face detection~\cite{liu2016joint}, 3D face reconstruction~\cite{chen2014joint}, and facial action unit activation detection~\cite{wu2016constrained}.

In the deep learning domain, coarse-to-fine approaches refine a coarse estimate of keypoints through a cascade~\cite{sun2013deep,zhou2013extensive,zhang2014coarse,zhang2016joint} or with branched networks~\cite{liang2015unconstrained}. Others assist keypoint detection by using separate cluster specific networks~\cite{wu2015facial}, augmenting it with related auxiliary tasks~\cite{zhang2014facial}, initializing with head pose predictions~\cite{yang2015face}, correcting for deformations with a spatial transformer~\cite{chen2016supervised}, incorporating shape basis and thin plate spline transformations~\cite{yu2016deep}, formulating keypoint detection as a dense 3D face model fitting problem~\cite{jourabloo2016large,zhu2016face}, or using deep regression models in combination with de-corrupt autoencoders~\cite{zhang2016occlusion}. Recent work explore using recurrent neural networks~\cite{peng2016recurrent,xiao2016robust,trigeorgis2016mnemonic}.

While deep learning approaches demonstrate impressive performance, they typically require large annotated datasets. Rather than collect a large dataset, \cite{masi2016we} uses domain specific augmentation techniques to \emph{synthesize} pose, shape, and expression variations. However, it relies on the availability of 3D face models, and addresses the related but separate problem of face recognition. Similarly, \cite{ding2016facenet2expnet} leverages large datasets available for face recognition to train a deep network, which is then used to guide training of an expression recognition network using only a small amount of data. However, while \cite{ding2016facenet2expnet} transfers knowledge between two different tasks (face recognition and expression recognition) that rely on the same type of data (human faces), we transfer knowledge between two different data sources (human and animal faces) in order to solve the same task (facial keypoint detection).

To the best of our knowledge, \emph{facial} keypoint detection in animals is a relatively unexplored problem. Very recently, \cite{yang2015human} proposed an algorithm for keypoint detection in sheep, using triplet interpolated features in a cascaded shape regression framework. Unlike our approach, it relies on hand-crafted features and does not transfer knowledge from human to animal faces. Keypoint localization on birds has been explored in \cite{singh2016learning,shih2015part,liu2014part,liu2013bird}, though these approaches do not focus on facial keypoint detection. %
\section{Approach}

Our goal is to detect facial keypoints in animals without the aid of a large annotated animal dataset.  To this end, we propose to adapt a pre-trained \emph{human} facial keypoint detector to \emph{animals} while accounting for their interspecies domain differences.  For training, we assume access to keypoint annotated animal faces, and keypoint annotated human faces and their corresponding pre-trained human keypoint detector.  For testing, we assume access to an animal face detector (i.e., we focus only on facial keypoint detection and not face detection).

Our approach has three main steps: (1) finding nearest neighbor human faces that have similar pose to each animal face; (2) using the nearest neighbors to train an animal-to-human warping network; and (3) using the warped (human-like) animal images to fine-tune a pre-trained human keypoint detector for animal facial keypoint detection.

\begin{figure}[t]
    \centering
    \includegraphics[width=0.48\textwidth]{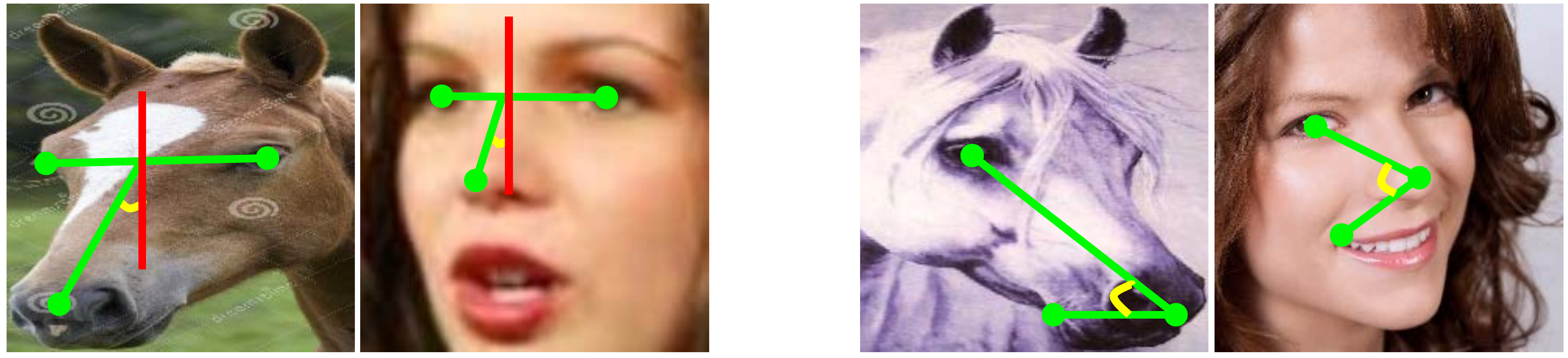}
    \caption{We approximate facial pose using the angle generated from the keypoint annotations.  The keypoints used to compute the angle-of-interest depend on which facial parts are visible.  For example, on the right, the horse's right eye and right mouth corner are not visible, so the three keypoints used are the left eye, nose, and left mouth corner.  While simple, we find this approach to produce reliable pose estimates.}
    \label{knn_fig}
\end{figure}

\subsection{Nearest neighbors with pose matching}

In order to fine-tune a (loosely-related) human facial keypoint detector to animals, our idea is to first warp the animal faces to have a more human-like shape so that it will be easier for the pre-trained human detector to adapt to the animal data.  One challenge is that an arbitrary animal and human face pair can exhibit drastically different poses (e.g., a right-facing horse and a left-facing person), which can making warping extremely challenging or even impossible.  To alleviate this difficulty, we first find animals and humans that are in similar poses.

If we had pose classifiers/annotations for both animal and human faces, then we could simply use their classifications/annotations to find compatible animal and human pairs.  However, in this work, we assume we do not have access to pose classifiers nor pose annotations.  Instead, we \emph{approximate} a face pose given its keypoint annotations.  More specifically, we compute the angular difference between a pair of human and animal keypoints, and then pick the nearest human faces for each animal instance.

\ignore{Formally, }For each animal training instance $A_i$, we find its nearest human neighbor training instance $H_{j*}$ based on pose:
\begin{equation}
nn(A_i)=H_{j*}=\argmin_{H_j} |\measuredangle^{*} A_i - \measuredangle^{*} H_j |,
\end{equation}
where $j$ indexes the entire human face training dataset, and the angle of interest $\measuredangle^{*}$ is measured in two different ways depending on the animal face's visible keypoints.  When both eyes and the nose are present, we use $\measuredangle^{*} = \measuredangle N E_{c} V$, where $E_{c}$ is the midpoint between the eye centers, $N$ is the nose position, and $V$ is a vertical line centered at $E_{c}$. If only the left eye is visible, then we use the left eye, nose, and left mouth keypoints: $\measuredangle^{*} = \measuredangle E_l N M_l$ (and $\measuredangle E_r N M_r$ if the right eye is visible).  These cases are illustrated in Fig.~\ref{knn_fig}.

\begin{figure}[t]
    \centering
    \includegraphics[width=0.485\textwidth]{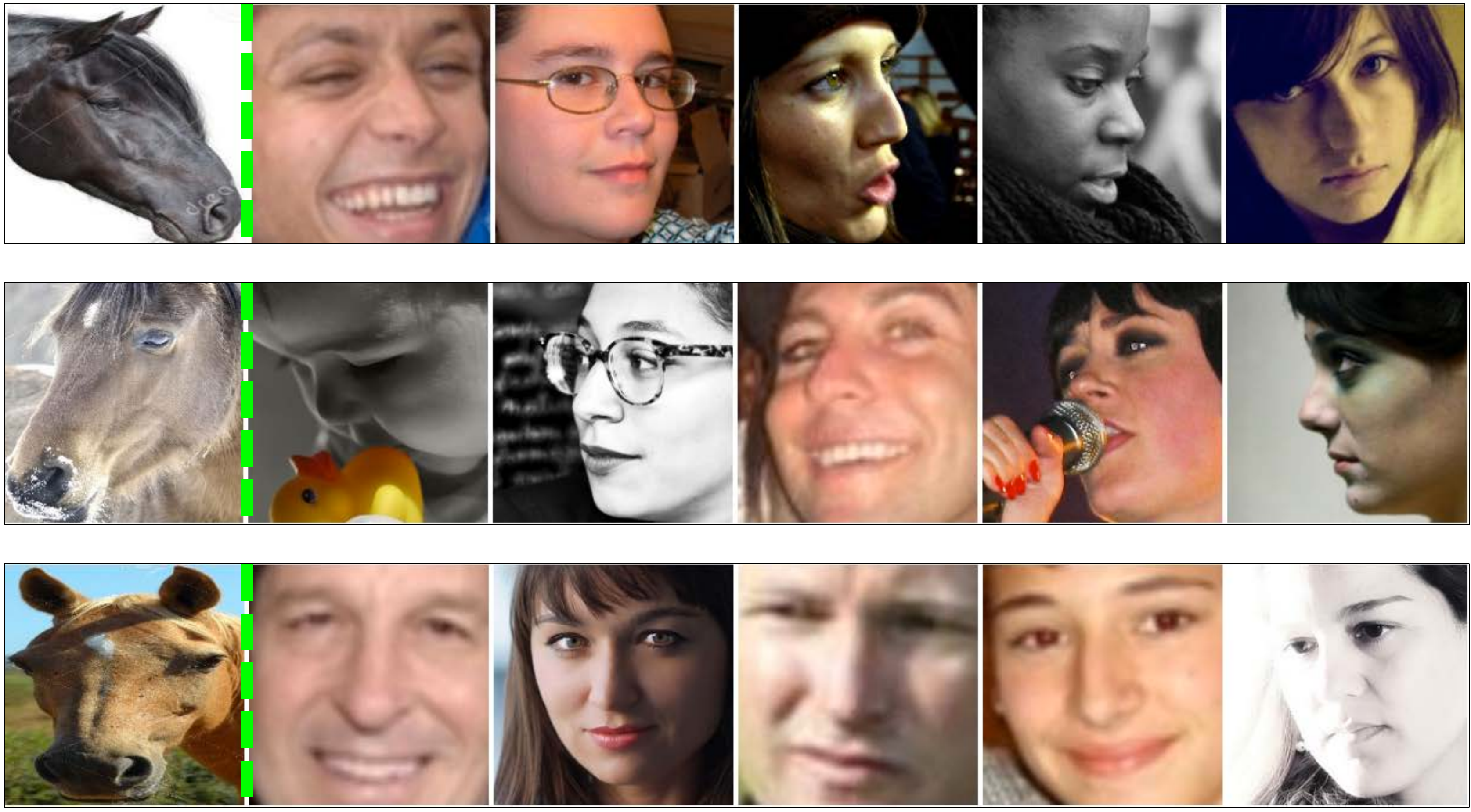}
    \caption{For each animal image (1st column), we find the nearest human neighbors in terms of pose.  These human neighbors are used to train a warp network that warps an animal to have human-like face shape.}
    \label{nn_fig}
\end{figure}

While simple, we find this approach to produce reliable pose estimates.  In our experiments, we find the $K=5$ nearest human neighbors for each animal face.  Fig.~\ref{nn_fig} shows some examples.  Since we use the TPS transformation for warping animals to humans (as described in the next section), we only compute matches for animal faces with at least three keypoints and ignore human matches whose keypoints are close to colinear, which can cause gross artifacts in warping.  Note that we do not do pose matching during testing, since we do not have access to ground-truth keypoints; instead we rely on the ensuing warping network to have learned the ``right'' warp for each animal face pose (based on its appearance) during training.

\begin{figure*}[t]
	\includegraphics[width=1\textwidth]{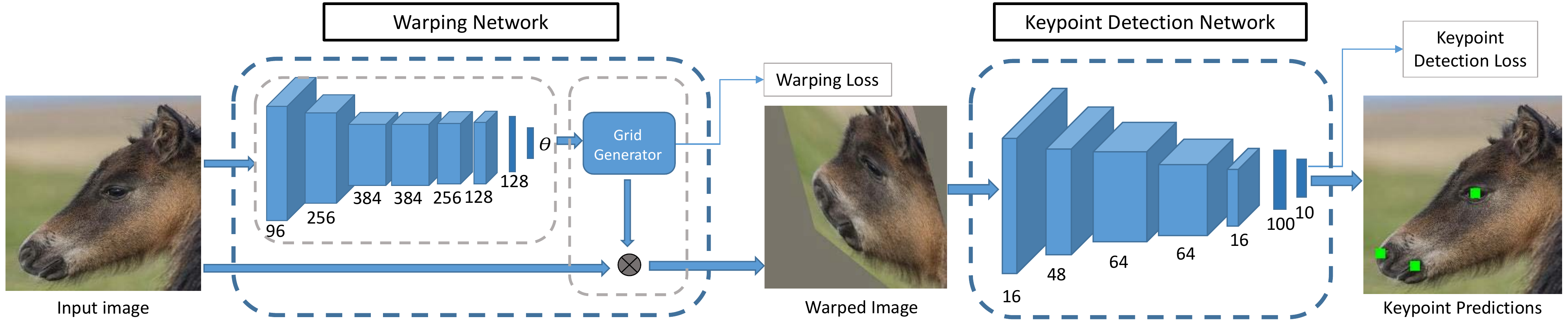}
	\caption{Our network architecture for animal facial keypoint detection.  During training, the input image is fed into the warping network, which is directly supervised using keypoint-annotated human and animal image pairs with similar pose.  The warping network warps the input animal image to have a human-like shape.  The warped animal face is then passed onto the keypoint detection network, which finetunes a pre-trained human keypoint detection network with the warped animal images. During testing, the network takes the input image and produces 5 keypoint predictions for left eye, right eye, nose, left mouth corner, and right mouth corner.}
	\label{fig:full_system}
\end{figure*}

\subsection{Interspecies face warping network}

Now that we have the nearest human faces (in terms of pose) for each animal face, we can use these matches to train an animal-to-human face warping network.  This warping network serves to adapt the shape of the animal faces to more closely resemble that of humans, so that a pre-trained human facial keypoint detector can be more easily fine-tuned on animal faces.

For this, we train a CNN that takes as input an animal image and warps it via a thin plate spline (TPS)~\cite{tps} transformation.  Our warping network is a spatial transformer~\cite{jaderberg2015spatial}, with the key difference being that our warps are directly supervised, similar to~\cite{chen2016supervised}.\footnote{In contrast, in~\cite{jaderberg2015spatial} the supervision only comes from the final recognition objective e.g., keypoint detection.  We show in Sec.~\ref{experiments} that direct warping supervision produces superior performance.}  Our network architecture is similar to the localization network in~\cite{krishna-eccv2016}; it is identical to Alexnet~\cite{krizhevsky-nips2012} up to the fifth convolutional layer, followed by a $1\times1$ convolution layer that halves the number of filters, two fully-connected layers, and batch normalization before every layer after the fifth. During training, the first five layers are pre-trained on ImageNet. We find these layer/filter choices to enable good TPS transformation learning without overfitting.  See Fig.~\ref{fig:full_system} (left).

For each animal and human training image pair, we first calculate the ground-truth TPS transformation using its corresponding keypoint pairs and apply the transformation to produce a ground-truth warped animal image.  We then use our warping network to compute a predicted warped animal image.  To train the network, we regress on the difference between the ground-truth warped image and predicted warped image pixel position offsets, similar to~\cite{kanazawa2016warpnet}.  Specifically, we use the squared loss to train the network:
\begin{equation}\label{equation_l2_squared}
	L_{warp}(A_i) = \sum_{m} (p_{i,m}^{pred}-p_{i,m}^{gt})^2,
\end{equation}
where $A_i$ is the $i$-th animal image, $p_{i,m}^{pred}$ and $p_{i,m}^{gt}$ are the predicted offset and ground-truth offset, respectively, for pixel $m$.

It is important to note that our warping network requires no additional annotation for training, since we only use the animal/human keypoint annotations to find matches (which are already available and necessary for training their respective keypoint detectors). In addition, since each animal instance has multiple ($K=5$) human matches, the warping network is trained to identify multiple transformations as potentially correct.  This serves as a form of data augmentation, and helps make the network less sensitive to outlier matches.

\subsection{Animal keypoint detection network}

Our warping network from the previous section conditions the distribution of the animal data to more closely resemble human data, so that we can harness the large \emph{human} keypoint annotated datasets that are readily available for \emph{animal} keypoint detection. The final step is to finetune a pre-trained human facial keypoint detection network to detect facial keypoints on our warped animal faces.

Our keypoint detector is a variant of the Vanilla CNN architecture used in~\cite{wu2015facial}. The network has four convolutional layers, and two fully-connected layers with absolute $\tanh$ non-linearity, and max-pooling in the last three convolutional layers.  We adapt it to work for larger images---we use $224\times224$ images as input rather than $40\times40$ used in~\cite{wu2015facial}---by adding an extra convolutional and max-pooling layer.  In addition, we add batch normalization after every layer since we find the $\tanh$ layers in the original network to be prone to saturation.\ignore{ We find batch normalization to also help the network adapt to smaller amounts of training data more easily.} Fig.~\ref{fig:full_system} (right) shows the architecture.  Our keypoint detection network is pre-trained on human facial keypoints on the AFLW~\cite{koestinger2011annotated} dataset and the training data used in~\cite{sun2013deep} (a total of 31524 images).

To\ignore{ train, and} finetune our keypoint network, we use the smooth $L1$ loss (equivalent to the Huber loss with $\delta$=1) used in \cite{girshick2015fast} since it is less sensitive to outliers that may occur with unusual animal poses:
\begin{equation}\label{equation_l1_smooth}
L_{keypoint}(A_i)= \sum_{n} smooth_{L_1}(k^{pred}_{i,n}-k^{gt}_{i,n}),
\end{equation}
where $A_i$ is the $i$-th animal image, $k^{pred}_{i,n}$ and $k^{gt}_{i,n}$ are the predicted and ground-truth keypoint position, respectively, for the $n$-th keypoint, and $smooth_{L_1}$ is
\begin{equation}
smooth_{L_1}(x)=\begin{cases}0.5x^2,& \text{if } |x|< 1\\
|x|-0.5,              & \text{otherwise.}
\end{cases}
\end{equation}

We set the loss for predicted keypoints with no corresponding ground-truth annotation (due to occlusion) to zero.

\subsection{Final architecture}

In our final model, we fit the warping network before a keypoint detection network that is pre-trained on human keypoint detection.  We use the two losses to jointly finetune both networks.  The keypoint detection loss $L_{keypoint}$ (Eqn.~\ref{equation_l1_smooth}) is back propagated through both the keypoint detection network, as well as the warping network. Additionally, the warping loss $L_{warp}$ (Eqn.~\ref{equation_l2_squared}) is backpropagated through the warping network, and the gradients are accumulated before the weights for both networks are updated.  See Fig.~\ref{fig:full_system}.

In the testing phase, our keypoint network predicts all 5 facial keypoints for every image.  In our experiments, we do not penalize the network for keypoint predictions that are not visible in the image and results are reported only for predicted keypoints that have corresponding ground-truth annotation. For evaluation,  the keypoints predicted on warped images are transferred back to the original image using the TPS warp parameters. \ignore{  In future work, it would be interesting to predict whether a keypoint is occluded.}

\subsection{Horse Facial Keypoint dataset}

As part of this work, we created a new horse dataset to train and evaluate facial keypoint detection algorithms.  We collected images through Google and Flickr by querying for ``horse face", ``horse head", and ``horse". In addition, we included images from the PASCAL VOC 2012~\cite{pascal-voc-2012} and Imagenet 2012~\cite{ILSVRC15} datasets.  There are a total of $3717$ images in the dataset: $3531$ for training, and $186$ for testing.  We annotated each image with face bounding boxes, and 5 keypoints: left eye center, right eye center, nose, left mouth corner, and right mouth corner.

\section{Experiments}\label{experiments}

In this section, we analyze our model's keypoint detection accuracy, and perform ablation studies to measure the contribution of each component.  In addition, we evaluate our method's performance as the amount of training data is varied, and also measure an upper-bound performance if animal-to-human warping were perfect.

\vspace{-10pt}
\paragraph{Baselines.} We compare against the algorithm presented in~\cite{yang2015human}, which uses triplet-interpolated features (TIF) in a cascaded shape regression framework for keypoint detection on animals.  We also develop our own baselines.  The first baseline is our full model without the warping network.  It simply finetunes the pre-trained human facial keypoint network on the animal dataset (``BL FT").  The second baseline is our full model without the warping loss; i.e., it finetunes the pre-trained human facial keypoint network and the warping network with only the keypoint detection loss.  This baseline is equivalent to the spatial transformer setting presented in~\cite{jaderberg2015spatial}.  We show results for this with TPS warps (``BL TPS"). The third baseline trains the keypoint detection network from scratch; i.e., without any human facial keypoint detection pretraining and without the warping network (``Scratch'').

\vspace{-10pt}
\paragraph{Datasets.} We pretrain our keypoint detection network on human facial keypoints from the AFLW~\cite{koestinger2011annotated} dataset and the training data used in~\cite{sun2013deep} (a total of 31524 images). This dataset is also used for animal to human nearest neighbor retrieval.  We evaluate keypoint detection on two animals: horses and sheep.  For the horse experiments, we use our Horse Facial Keypoint dataset, which consists of 3531 images for training and 186 for testing.  For the sheep experiments, we manually annotated a subset of the dataset provided in~\cite{yang2015human} with mouth corners so that we have the same 5 keypoints present in the human dataset.  The dataset consists of 432 images for training and 99 for testing.

\vspace{-10pt}
\paragraph{Evaluation metric.} We use the same metric for evaluation as~\cite{yang2015human}: If the euclidean distance between the predicted and ground-truth keypoint is more than 10\% of the face (bounding box) size, it is considered a failure.  We then compute the average failure rate as the percentage of testing keypoints that are failures.

\begin{figure}[t]
    \includegraphics[width=0.235\textwidth]{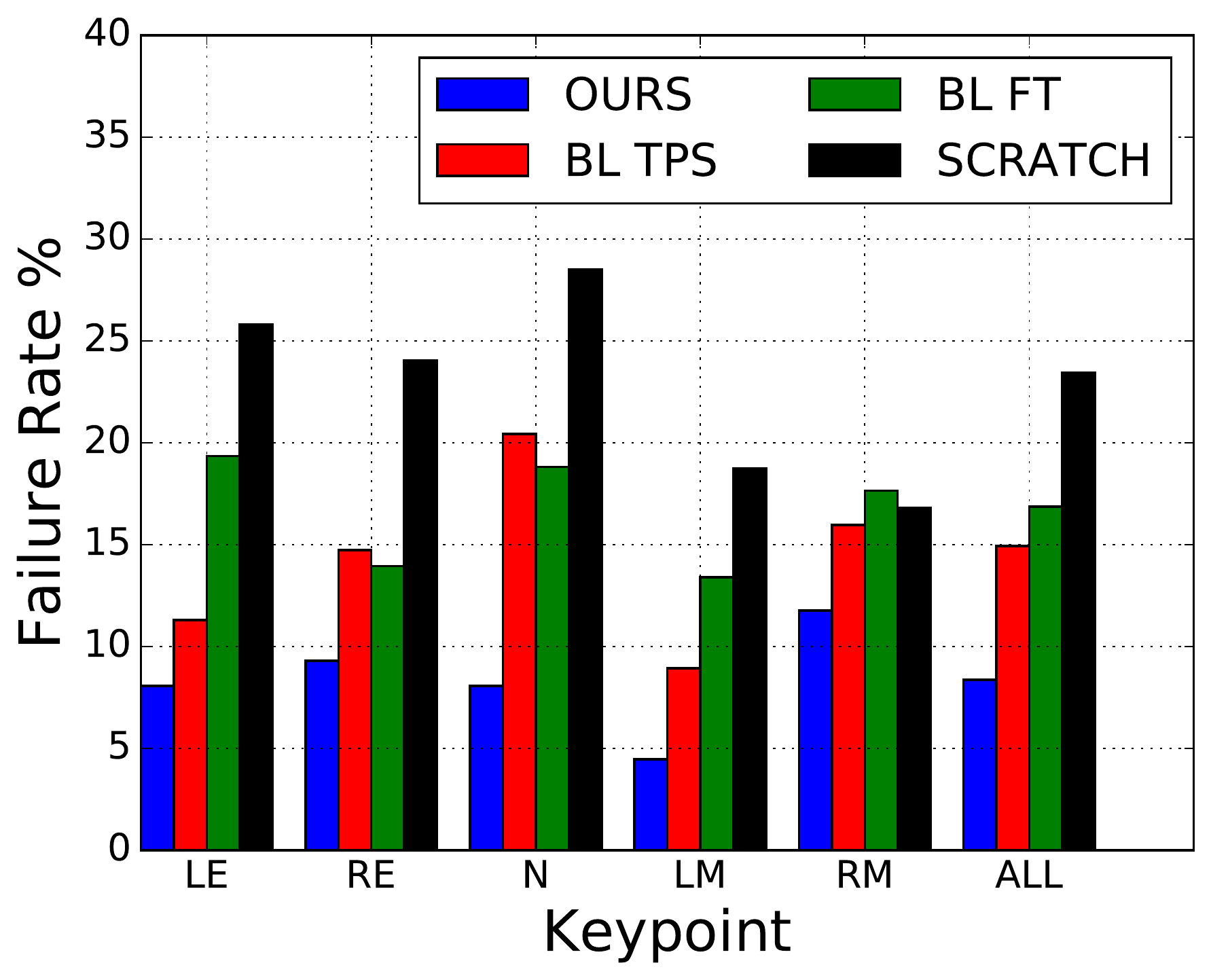}
	\includegraphics[width=0.235\textwidth]{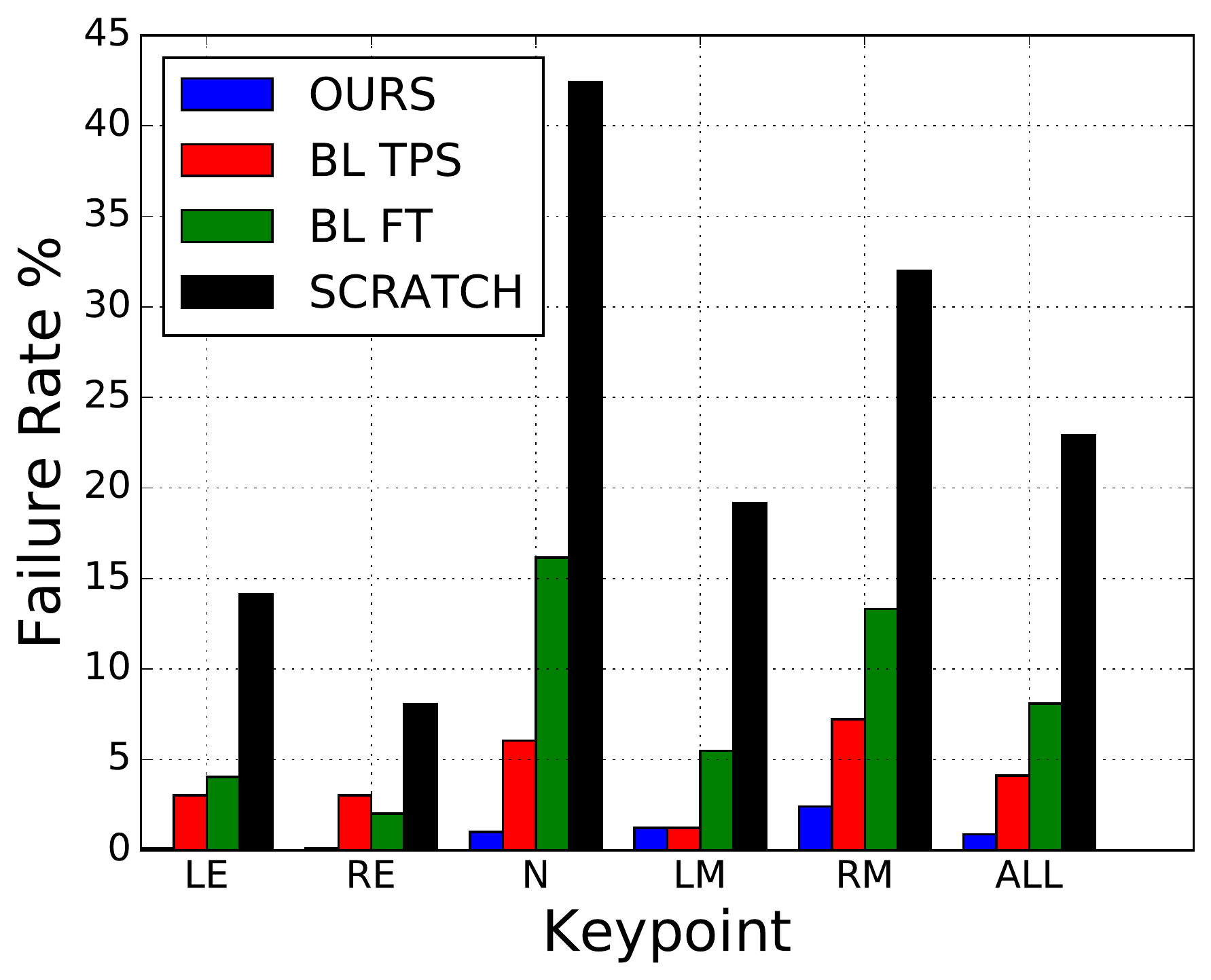}
	\caption{Average keypoint detection failure rate (\% of predicted keypoints whose euclidean distance to the corresponding ground-truth keypoint is more than 10\% of the face bounding box size).  Horses (\textbf{left}) and Sheep (\textbf{right}).  Our approach outperforms the baselines.  \emph{Lower is better.}  See text for details.}
	\label{baselines}
\end{figure}

\vspace{-10pt}
\paragraph{Training and implementation details.}
We find that pretraining the warping network before joint training leads to better performance.  To train the warping and keypoint network, we use $K=5$ human neighbors for each animal instance.\ignore{ During final training, we also use $K=5$ human neighbors to supervise the warping network. We found this helped prevent overfitting, and served as a form of data augmentation.} These matches are also used to supervise the ``GT Warp" network described in Sec.~\ref{sec:ablation}.

For the TPS warping network, we use a $5\times5$ grid of control points. We optimize all networks using Adam~\cite{Kingma-arxiv2014}. The base learning rate for the warp network training is 0.001, with a $\frac{1}{10}$$\times$ lower learning rate for the pre-trained layers.  It is trained for 50 epochs, with the learning rate lowered by $\frac{1}{10}$$\times$ after 25 epochs.  During full system training, the warp network has the same learning rates, while the keypoint detection network has a learning rate of 0.01.  We train the network for 150 epochs, lowering the learning rate twice after 50 and 100 epochs.  Finally, we use horizontal flips and rotations from $-10\degree$ to $10\degree$ at increments of $5\degree$ for data augmentation.

\begin{figure*}[t]
	\centering
	\begin{tabular}{ cccc }
	\includegraphics[width=0.23\textwidth]{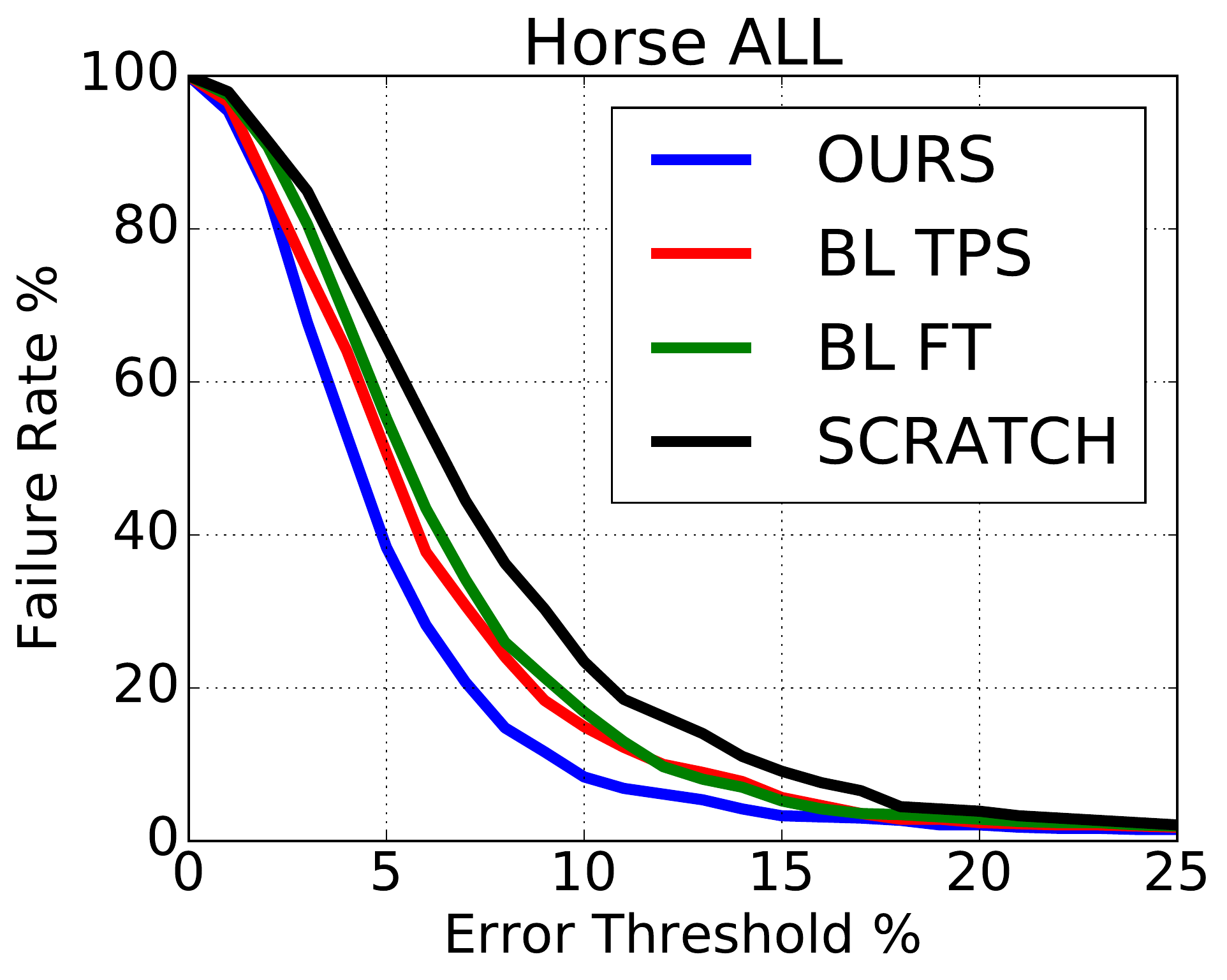} &
	\includegraphics[width=0.23\textwidth]{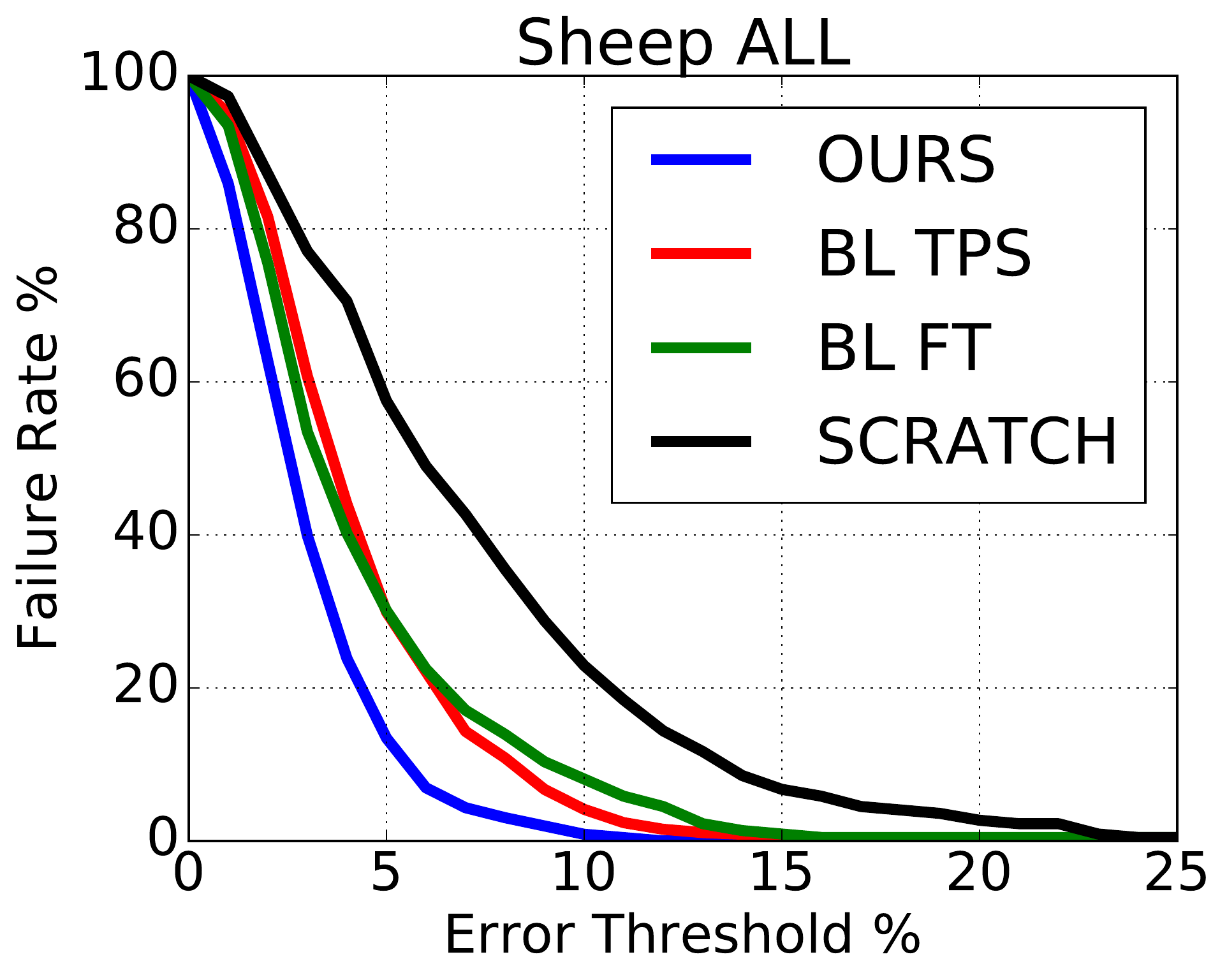} &
		\includegraphics[width=0.23\textwidth]{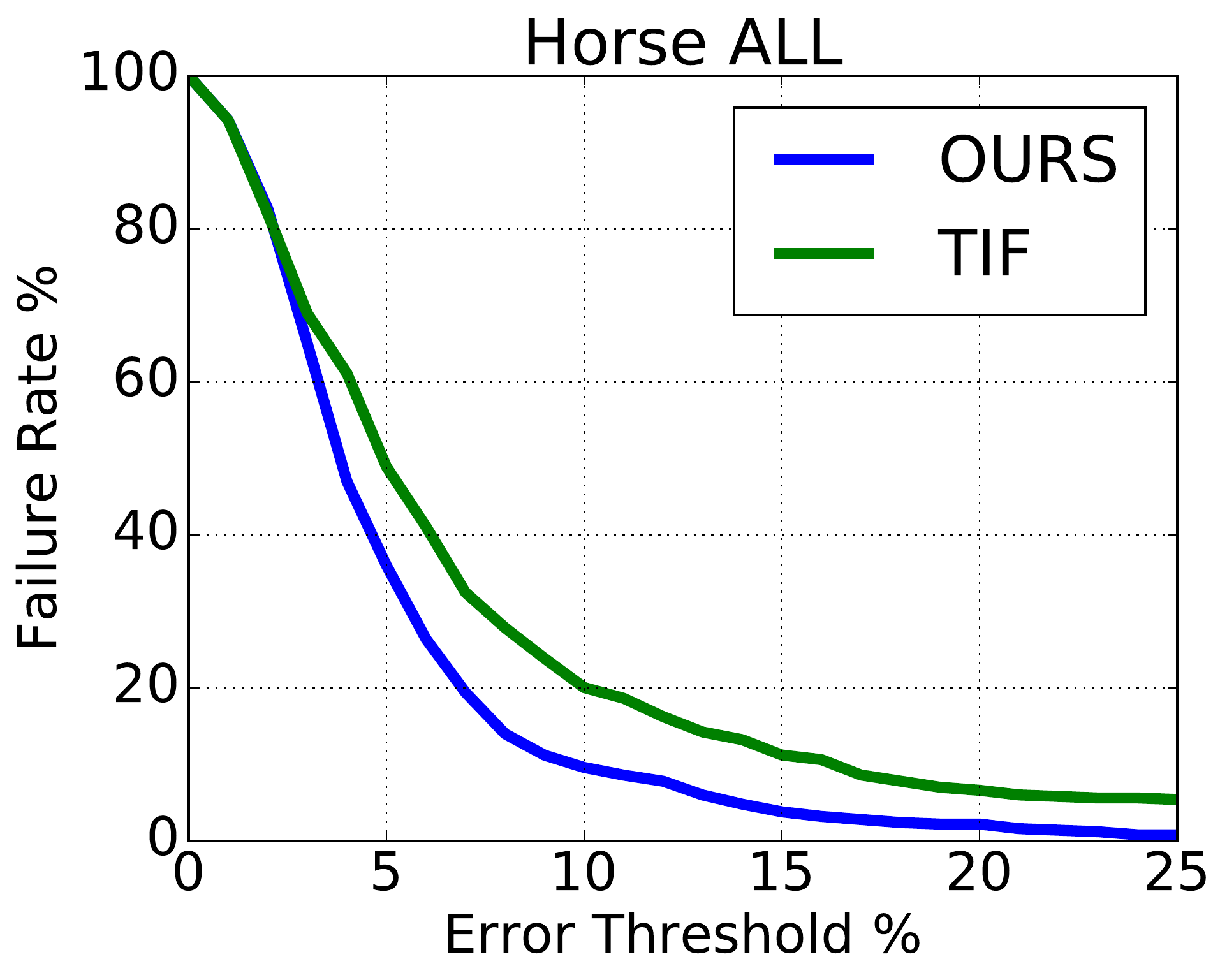} &
		\includegraphics[width=0.23\textwidth]{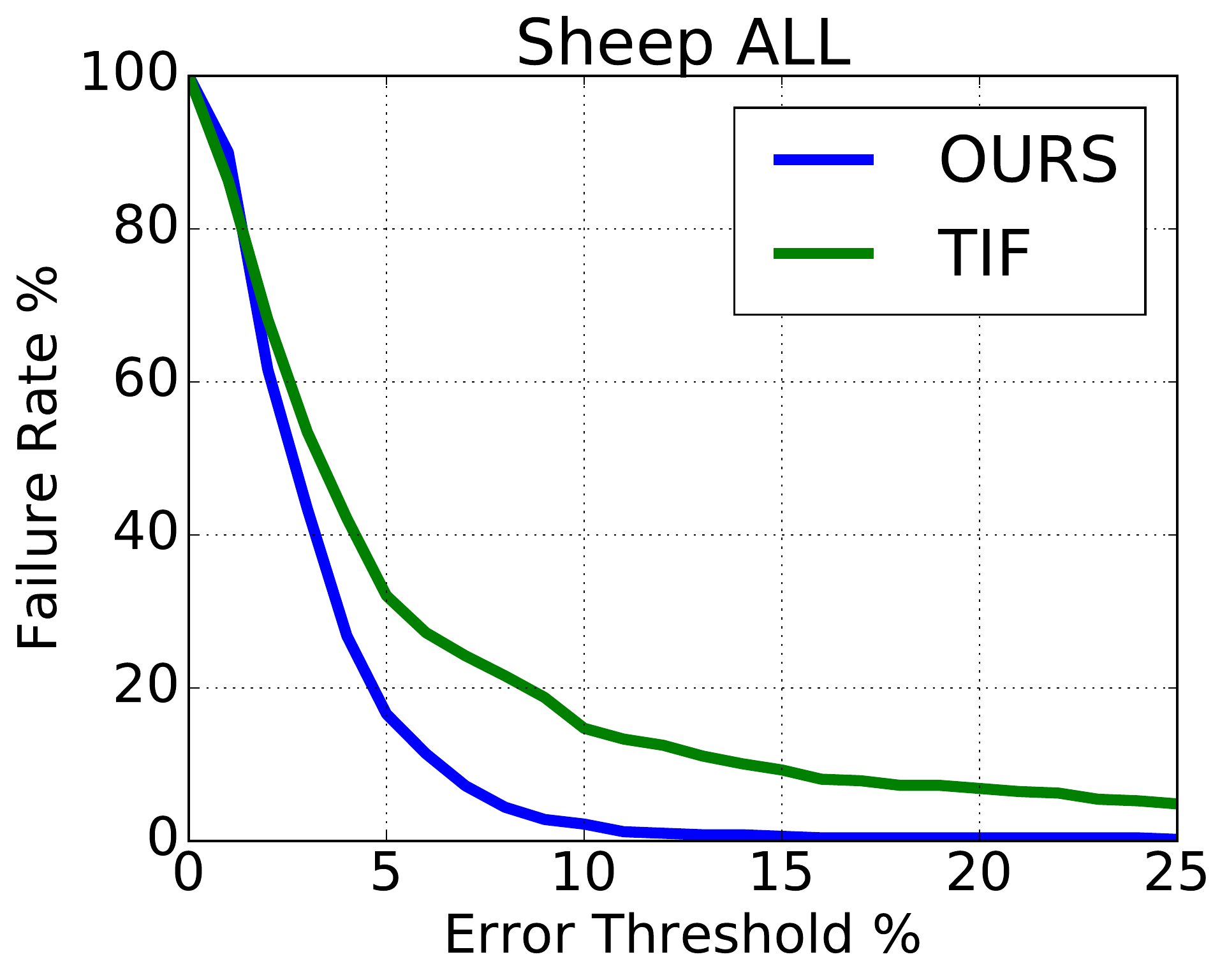}
	\end{tabular}
	\caption{Average keypoint detection failure rate across all keypoints for our system vs.~our baselines (first two plots) and the Triplet Interpolated Features (TIF) approach of Yang et al.~\cite{yang2015human} (last two plots). Our system sustains lower failure rates across stricter failure thresholds than all baselines.}
	\label{fig_vary_thresh}
\end{figure*}

\begin{figure}[t]
	\includegraphics[width=0.235\textwidth]{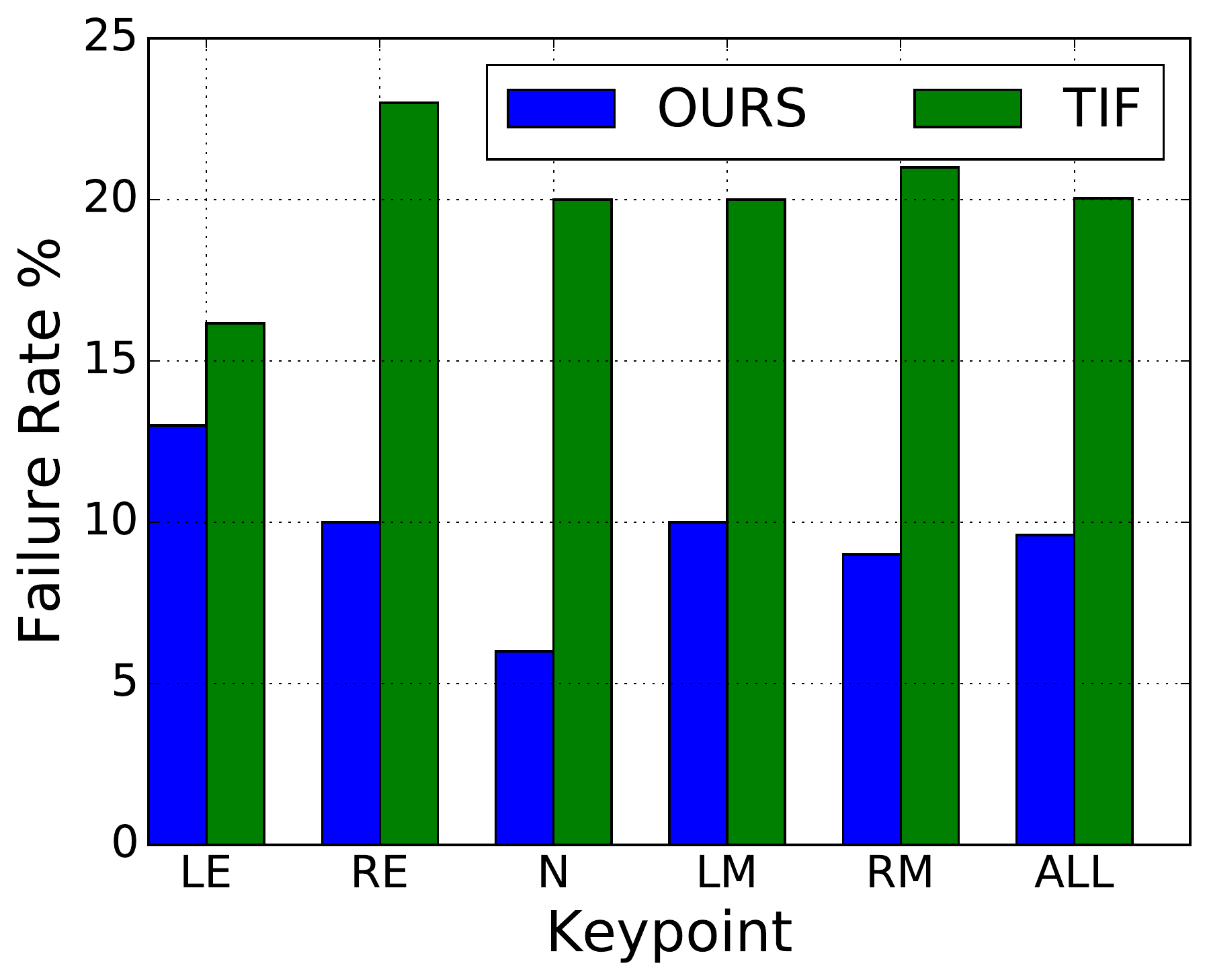}
	\includegraphics[width=0.235\textwidth]{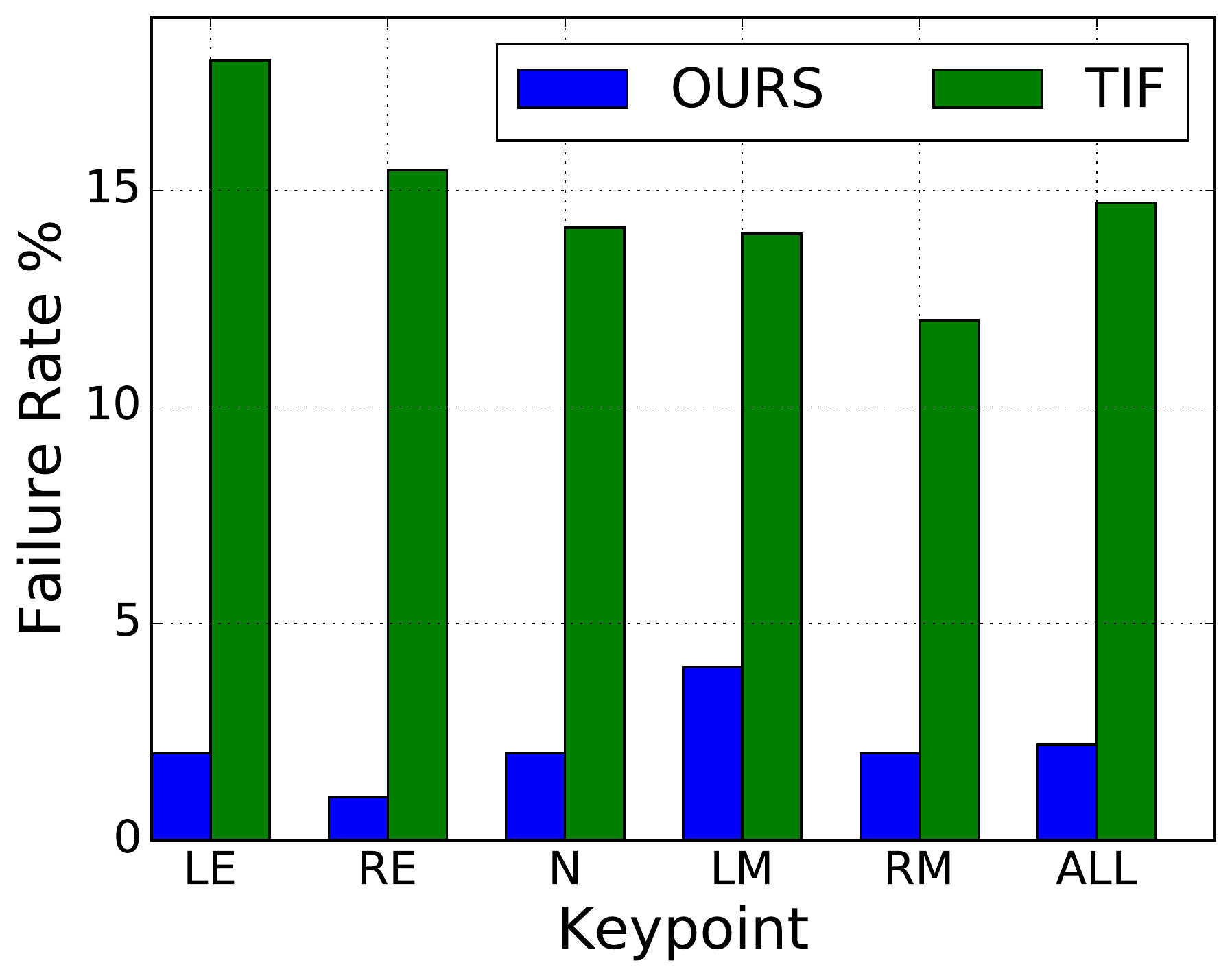}
	\caption{Average keypoint detection failure rate for Horses (\textbf{left}) and Sheep (\textbf{right}).  Our approach significantly outperforms the Triplet Interpolated Features (TIF) approach of Yang et al.~\cite{yang2015human}, which combines hand-crafted features with cascaded shape regressors.  \emph{Lower is better.}}
	\label{tif}
\end{figure}

\subsection{Comparison with our baselines}

We first compare our full model with our model variant baselines.  Figure~\ref{baselines} (left) and (right) show results on horse and sheep data, respectively.  We outperform all of our baselines significantly for both horses and sheep, with an average failure rate across keypoints at 8.36\% and 0.87\%, respectively.%

Overall, the failure rate for all methods (except Scratch) for sheep is lower than that for horses. The main reason is due to the pose distribution of human and sheep data being more similar than that of human and horse data. The human and sheep data have 72\% and 84\% of images in frontal pose (faces with all 5 keypoints visible) as compared to only 29\% for horses.  The majority (60\%) of horse faces are side-view (faces with only 3 keypoints visible). This similarity makes it easier for the human pre-trained network to adapt to sheep than to horses. Nonetheless, the fact that our method outperforms the baselines for both datasets, demonstrates that our idea is generalizable across different types of data.

These results also show the importance of each component of our system. Training with a human pre-trained network does better than training from scratch (BL FT vs.~Scratch); adding a warping network that is only \emph{weakly-guided} by the keypoint detection loss further improves results (BL TPS vs.~BL FT); and finally, directly supervising the warping network to produce animal faces that look more human-like leads to the best performance (Ours vs.~BL TPS). The first two plots in Fig.~\ref{fig_vary_thresh} show the results of varying the acceptance threshold (on the euclidean distance between the ground-truth and predicted keypoint) for a valid keypoint on our and the baselines' performance. Our method sustains superior accuracy across thresholds, which again indicates that we predict keypoints more accurately.

\begin{figure}[t]
    \centering
	\includegraphics[width=0.48\textwidth]{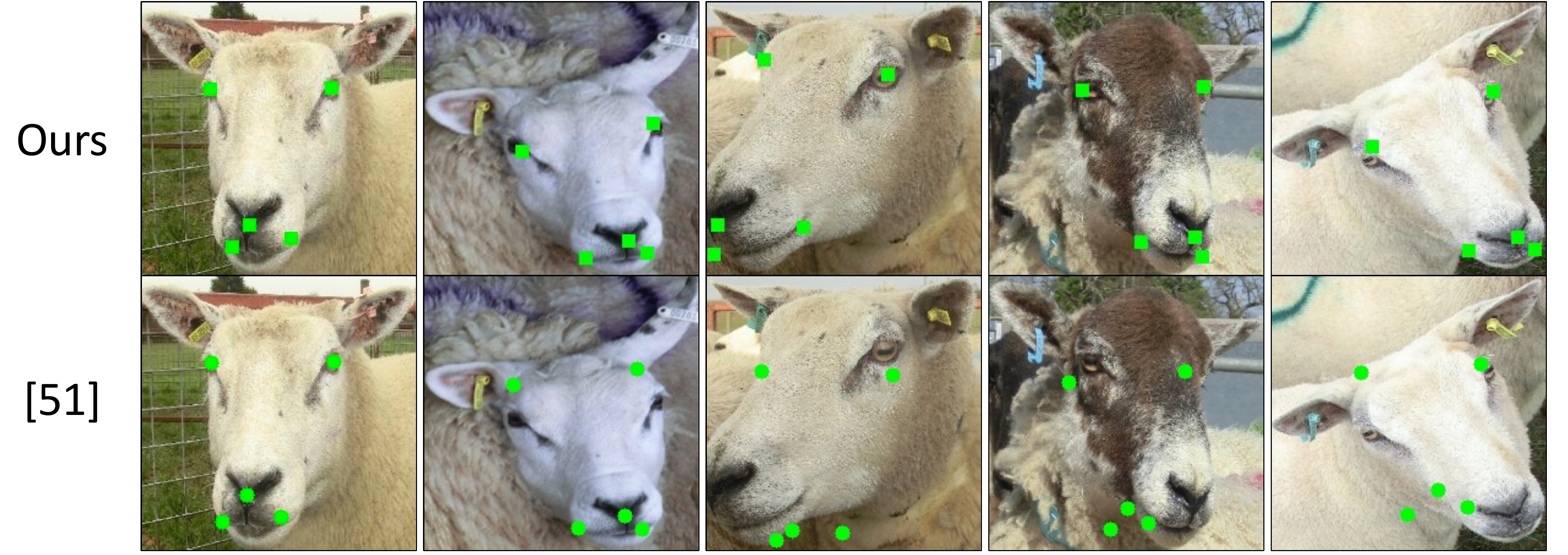}
	\caption{Qualitative examples comparing our approach and Yang et al.~\cite{yang2015human} on their Sheep dataset.  While \cite{yang2015human} can produce good predictions (first column), overall, our method produces significantly more accurate results.}
	\label{fig_sheep}
\end{figure}

\begin{figure*}[t]
    \centering
	\includegraphics[width=0.9\textwidth]{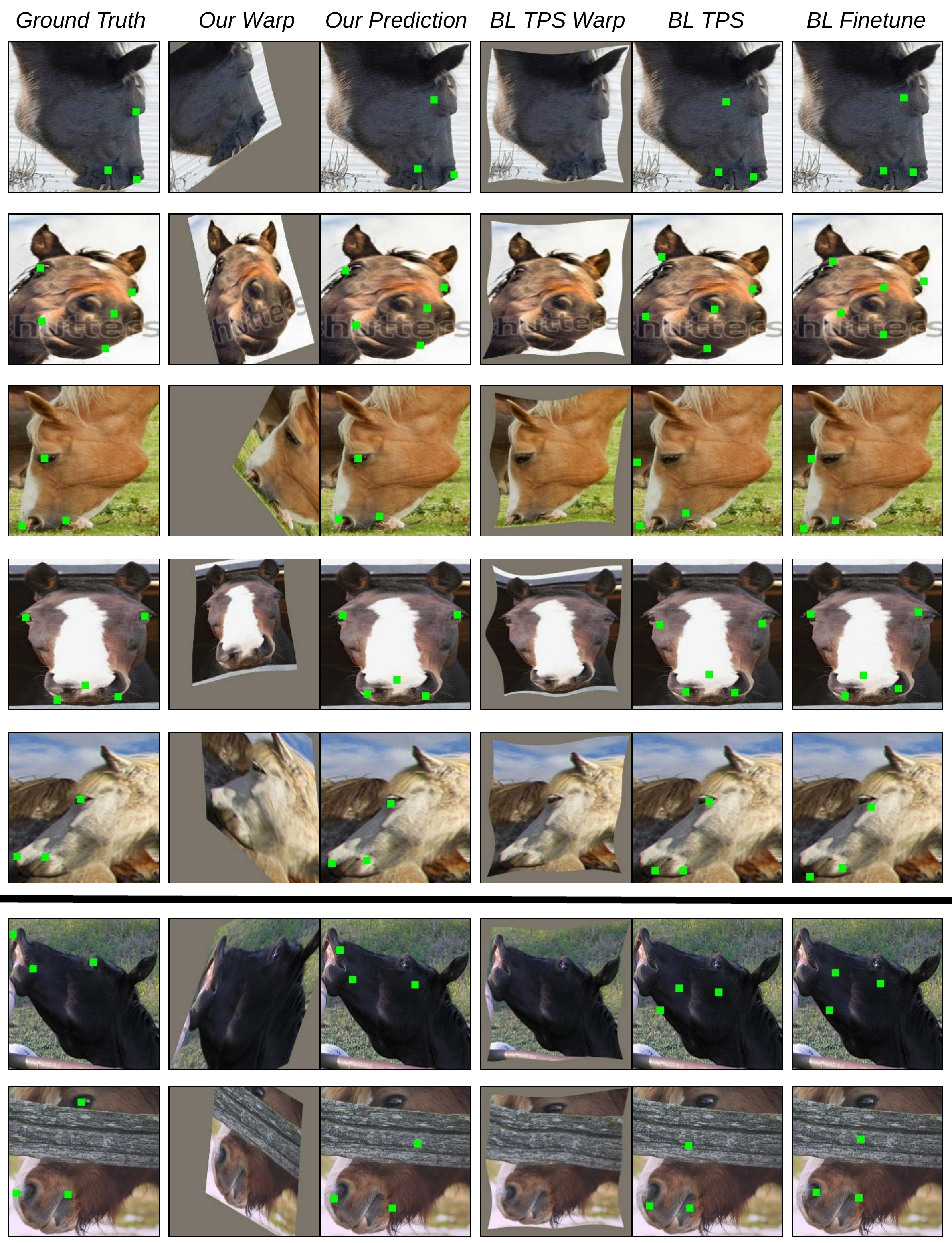}
	\caption{Qualitative examples of predicted keypoints and predicted warps for ours and the baselines.  The first five rows show examples where our method outperforms the baseline.  While the baselines also produce reasonable results, by warping the horses to have more human-like shape, our method produces more precise keypoint predictions. For example, in the first row, the baselines do not localize the nose and mouth corner as well as ours.  The last two rows show typical failure examples due to extreme pose or occlusion.}
	\label{fig_horse}
\end{figure*}

Fig.~\ref{fig_horse} shows qualitative examples of predicted keypoints and predicted warps for ours and the baselines. Noticeably, the TPS warps produced without the warping loss (BL TPS Warp) fail to distinguish between the different horse poses, and also do not warp the horse faces to look more human like.\ignore{ Noticeably, our warps tend to be smoother than the TPS warps (produced without the warping loss), \yj{which can sometimes produce gross deformations}.}  \ignore{This is due to two main reasons: our direct supervision of the warping network, as well as use of multiple human nearest neighbors for matching, which regularizes the warp.}On the other hand, our warping network is able to do both tasks well since it is directly supervised by pose specific human matches. By warping the horses to have more human-like shape, our method produces more precise keypoint predictions than the baselines.  The last two rows show typical failure examples due to extreme pose or occlusion.

\subsection{Comparison with Yang et al.~\cite{yang2015human}}

We next compare our method to the Triplet Interpolated Features (TIF) approach of~\cite{yang2015human}, which is the state-of-the-art animal keypoint detector.  The method requires the existence of all landmarks in all training examples.  We therefore picked a subset of the horse and sheep images where all $5$ keypoints are visible and marked: 345/100 train/test images for sheep, and 982/100 train/test images for horses.

Fig.~\ref{fig_sheep} shows qualitative examples comparing our method's keypoint predictions vs.~those made by TIF.  TIF often fails to\ignore{ robustly} handle large appearance and pose variations.  This is also reflected in the quantitative results, which are shown in Fig.~\ref{fig_vary_thresh} (third) and Fig.~\ref{tif} (left) for the horse dataset and Fig.~\ref{fig_vary_thresh} (fourth) and Fig.~\ref{tif} (right) for the sheep dataset. We significantly outperform TIF on both datasets (10.44\% and 12.52\% points lower failure rate for horses and sheep, respectively).  The main reason is because we use a high capacity deep network, whereas TIF is a shallow method that learns with hand-crafted features.  Importantly, the reason that we are able to use such a high capacity deep network---despite the limited training data of the animal datasets---is precisely because we correct for the shape differences between animals and humans in order to \emph{finetune} a pre-trained human keypoint detection network.

\begin{figure}[t]
    \centering
    \hspace{-1pt}
	\includegraphics[width=0.22\textwidth]{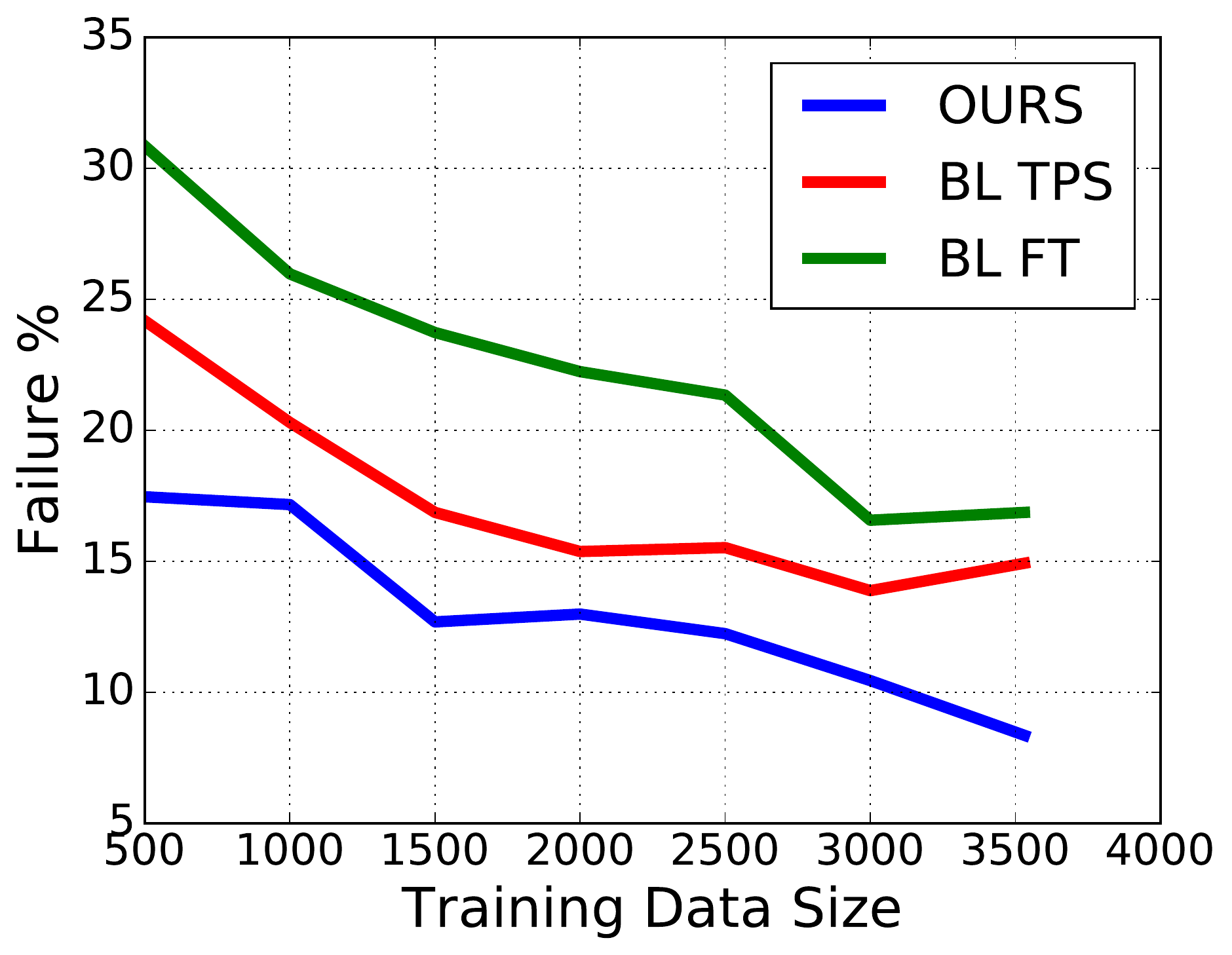}
	\includegraphics[width=0.51\linewidth]{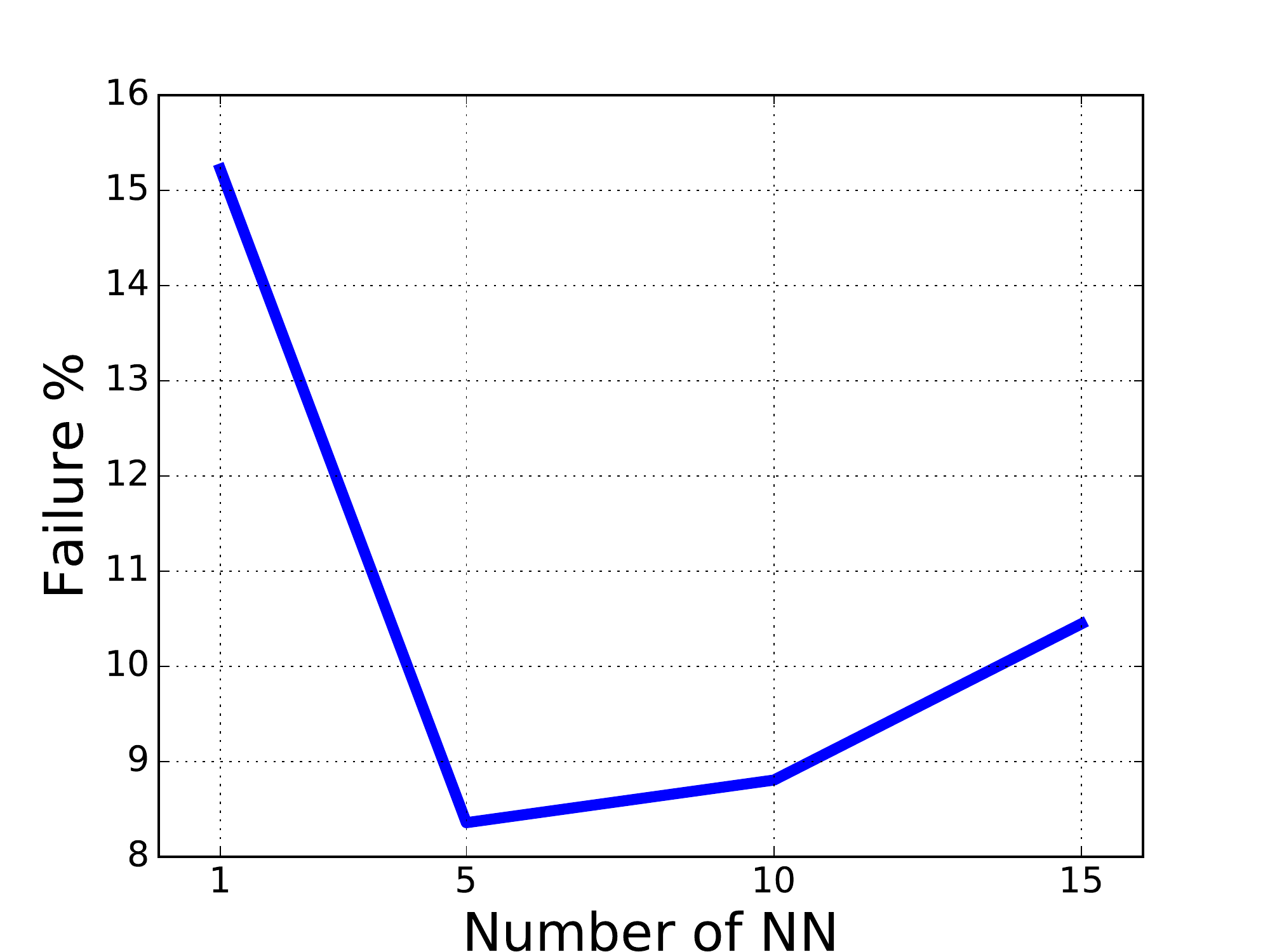}%
	\caption{(\textbf{left}) Average keypoint detection failure rate as a function of the number of training instances on the Horse dataset.  Our failure rate increases more gracefully compared to the baselines as the number of training images is decreased. \emph{Lower is better.}  (\textbf{right}) Increasing the number of human face neighbors for an animal face instance increases performance until noisy neighbors cause performance to drop.} %
	\label{fig_ablation_data}
\end{figure}

\subsection{Effect of training data size}

In this section, we evaluate how the performance of our network changes as the amount of training data varies.  For this, we train and test multiple versions of our model and the baselines, each time using 500 to 3531 training images in 500 image increments on the Horse dataset.

Figure~\ref{fig_ablation_data} (left) shows the result. While the performance of all methods decreases with the training data amount, our performance suffers much less than that of the simple finetuning and TPS baselines. In particular, when using only 500 training images, our method has a 6.72\% point lower failure rate than the TPS baseline while relying on the same network architecture, and a 13.39\% point lower failure rate than simple finetuning, \emph{without using any additional training data or annotations}.

This result demonstrates that our algorithm adapts well to small amounts of training data, and bolsters our original argument that explicitly correcting for interspecies shape differences enables better finetuning, since the pre-trained human keypoint detection network can mostly focus on the appearance differences between the two domains (humans and animals).  Importantly, it also shows the practical applicability of our approach to small datasets.

\subsection{Effect of warping accuracy}\label{sec:ablation}%

We next analyze the influence of warping accuracy on keypoint detection.  For this, we first analyze the performance of our keypoint detection network when finetuned with \emph{ground-truth} warped images (``GT Warp"), where we use the ground-truth keypoint annotations between human and horse faces for warping (i.e., the keypoint detection network is finetuned with ground-truth warped images).  In a sense, this represents the upper bound of the performance of our system.

\begin{table}[t!]
\centering
    \begin{tabular}{ |c||c|c| }
		\hline
		& GT Warp & Ours \\
		\hline
		Failure Rate \% & 7.76\%  & 8.36\% \\
		\hline
	\end{tabular}
    \caption{Average keypoint detection failure rate across all keypoints on the Horse dataset, comparing our approach to an upper-bound ground-truth warping baseline. \emph{Lower is better.}}
    \label{table_results}
    \vspace{-0.1in}
\end{table}

Table~\ref{table_results} shows the results on our Horse dataset.  First, the GT Warp upper-bound produces even lower error rates than our method\ignore{ (and all the baselines)}, which demonstrates the efficacy of the idea of correcting for shape differences by warping.  At the same time, the non-negligible error rate of GT Warp also hints at the limitation of our warping network's training data and/or pose matching strategy.  Better training data, with either a different algorithm for nearest pose neighbor matching or an increase in the keypoints that are annotated\ignore{ (i.e., beyond just five keypoints)} could potentially lead to a better upper-bound, and would likely provide improvements for our approach as well.

\subsection{Evaluation of Nearest Neighbors}
Finally, we evaluate the importance of human nearest neighbors for our system. We vary the number of nearest neighbors used for training our full system from $K=1$ to $K=15$ at increments of $5$ for our full Horse training set.  The result is shown in Figure~\ref{fig_ablation_data} (right).  While the error rate decreases as the number of neighbors used for training is increased in the beginning, eventually, the noise in retrieved nearest neighbors causes the error rate to increase.

\section{Conclusion}

We presented a novel approach for localizing facial keypoints on animals.  Modern deep learning methods typically require large annotated datasets, but collecting such datasets is a time consuming and expensive process.\ignore{ For context, the ethical research standards on Amazon Mechanical Turk recommend that each Turker earn between \$6-8 per hour if they work continuously~\cite{salehi2015we} -- assuming an annotation rate of one image per minute, it would take close to 2 years of annotation, and at least \$100,000 to collect a million image training dataset.}

Rather than collect a large annotated animal dataset, we instead warp an animal's face shape to look like that of a human.  In this way, our approach can harness the readily-available human facial keypoint annotated datasets for the loosely-related task of animal facial keypoint detection.  We compared our approach with several strong baselines, and demonstrated state-of-the-art results on horse and sheep facial keypoint detection.  Finally, we introduced a novel Horse Facial Keypoint dataset, which we hope the community will use for further research on this relatively unexplored topic of animal facial keypoint detection.

\vspace{-10pt}
\paragraph{Acknowledgements.} This work was supported in part by a gift from the Swedish University of Agricultural Sciences and GPUs donated by NVIDIA.

{\small
\bibliographystyle{ieee}
\bibliography{egbib,strings}
}

\end{document}